\documentclass[sigconf, nonacm]{acmart}
\usepackage{booktabs}
\usepackage{multirow}
\usepackage{siunitx}  
\usepackage{array}
\usepackage{bm}

\AtBeginDocument{
  }
    
\setcopyright{none}
\copyrightyear{2026}
\acmYear{2026}
\acmConference[]{}{}{}
\acmBooktitle{}
\acmDOI{}
\acmISBN{}

\begin{document}

\title{ProRAG: Process-Supervised Reinforcement Learning \\ for Retrieval-Augmented Generation}

\author{Zhao Wang}
\affiliation{Gaoling School of Artificial Intelligence, Renmin University of China
  \city{Beijing}
  \country{China}
}
\email{lilin22wz@gmail.com}
\orcid{0009-0004-7729-5668}

\author{Ziliang Zhao}
\affiliation{Gaoling School of Artificial Intelligence, Renmin University of China
  \city{Beijing}
  \country{China}
}
\email{zhaoziliang@ruc.edu.cn}
\orcid{0000-0001-8169-0143}

\author{Zhicheng Dou}
\authornote{Zhicheng Dou is the corresponding author.}
\affiliation{Gaoling School of Artificial Intelligence, Renmin University of China
  \city{Beijing}
  \country{China}
}
\email{dou@ruc.edu.cn}
\orcid{0000-0002-9781-948X}

\begin{abstract}
Reinforcement learning (RL) has become a promising paradigm for optimizing Retrieval-Augmented Generation (RAG) in complex reasoning tasks. However, traditional outcome-based RL approaches often suffer from reward sparsity and inefficient credit assignment, as coarse-grained scalar rewards fail to identify specific erroneous steps within long-horizon trajectories. This ambiguity frequently leads to ``process hallucinations'', where models reach correct answers through flawed logic or redundant retrieval steps. Although recent process-aware approaches attempt to mitigate this via static preference learning or heuristic reward shaping, they often lack the on-policy exploration capabilities required to decouple step-level credit from global outcomes. To address these challenges, we propose \textbf{ProRAG}, a process-supervised reinforcement learning framework designed to integrate learned step-level supervision into the online optimization loop. Our framework consists of four stages: (1) \textbf{Supervised Policy Warmup} to initialize the model with a structured reasoning format; (2) construction of an \textbf{MCTS-based Process Reward Model (PRM)} to quantify intermediate reasoning quality; (3) \textbf{PRM-Guided Reasoning Refinement} to align the policy with fine-grained process preferences; and (4) \textbf{Process-Supervised Reinforcement Learning} with a dual-granularity advantage mechanism. By aggregating step-level process rewards with global outcome signals, ProRAG provides precise feedback for every action. Extensive experiments on five multi-hop reasoning benchmarks demonstrate that ProRAG achieves superior overall performance compared to strong outcome-based and process-aware RL baselines, particularly on complex long-horizon tasks, validating the effectiveness of fine-grained process supervision. The code and model are available at \url{https://github.com/lilinwz/ProRAG}.
\end{abstract}



\maketitle

\section{Introduction}

Recent Large Language Models (LLMs) have demonstrated remarkable reasoning capabilities~\cite{gpt4, deepseek}, significantly improving their performance on complex, knowledge-intensive tasks. These advances have propelled Retrieval-Augmented Generation (RAG) from static pipelines into agentic systems, where models act as autonomous decision-makers that actively plan retrieval strategies, interact with external tools, and iteratively refine their reasoning based on feedback~\cite{searcho1,deeprag}. However, optimizing such dynamic workflows with standard supervised learning remains challenging. Effective reasoning and retrieval trajectories are often open-ended, which makes fine-grained ground-truth supervision for intermediate decisions unavailable or prohibitively expensive. To bridge this gap, Reinforcement Learning (RL) has been increasingly adopted to optimize the model policy from the final outcome signals~\cite{searchr1, r1searcher}.

\begin{figure}[t]
  \centering
  \includegraphics[width=0.8\columnwidth]{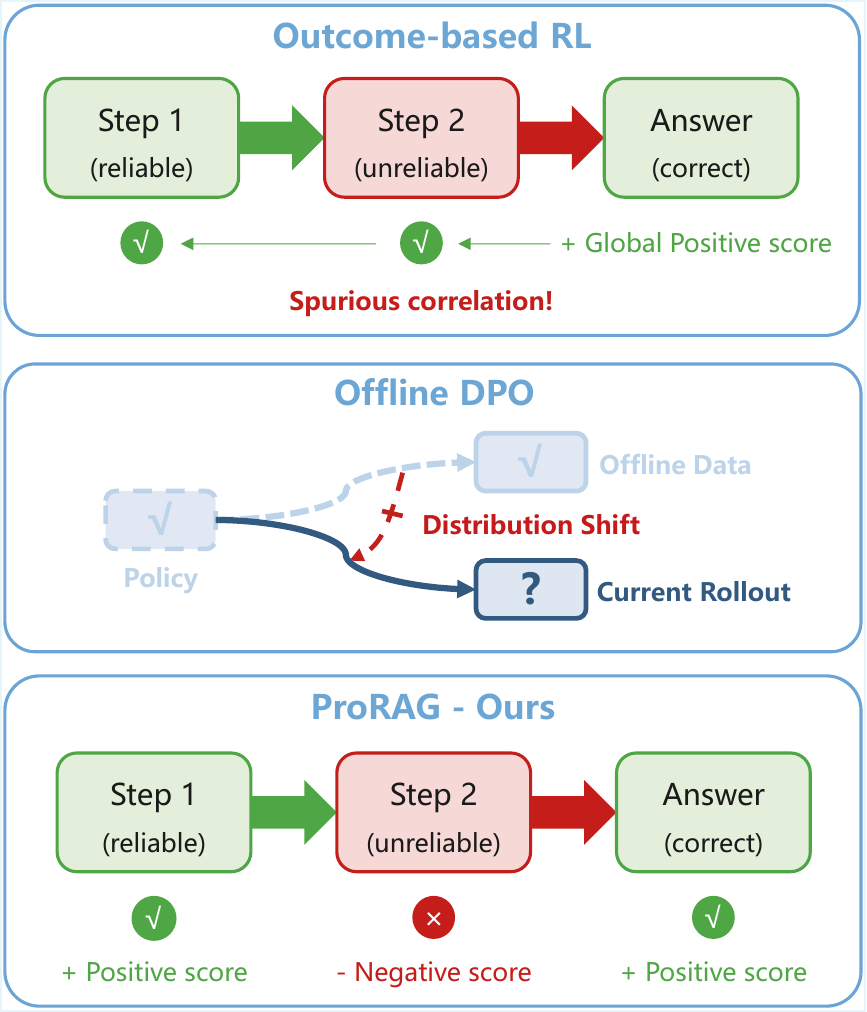}
  \caption{Comparison of optimization paradigms. \textbf{(a) Outcome-based RL} suffers from spurious correlations due to sparse global signals. \textbf{(b) Offline DPO} is limited by distribution shifts caused by static datasets. \textbf{(c) ProRAG (Ours)} utilizes dense step-level supervision for precise credit assignment and dynamic error correction.}
  \label{fig:comparison}
\end{figure}

However, traditional outcome-based reinforcement learning often suffers from reward sparsity and the credit assignment problem, making it ineffective for complex multi-hop RAG tasks. In such long-horizon trajectories, a coarse-grained scalar reward based on final answer correctness often fails to distinguish the specific contributions of intermediate actions. This ambiguity frequently results in ``process hallucinations'', where the model is rewarded for reaching the correct answer while relying on flawed reasoning, redundant retrieval, or unverified assumptions. Consequently, such outcome-oriented feedback can inadvertently reinforce spurious correlations rather than authentic reasoning abilities. Furthermore, without dense supervision to pinpoint specific errors, the model may struggle to correct early deviations during training, resulting in inefficient optimization.

To address these challenges, recent studies have explored optimization strategies for incorporating process-level signals, broadly categorized into offline and online approaches. Offline methods~\cite{reasonrag, decexrag, rearter} leverage search algorithms like MCTS to synthesize preference trajectories for Direct Preference Optimization (DPO), yet this reliance on static data constrains the model’s ability to adapt to on-policy distribution shifts. Conversely, online methods often resort to heuristic reward shaping~\cite{hiprag, tiresrag}, step-wise normalization~\cite{ppr, stepsearch}, or multi-agent interaction~\cite{sirag} to stabilize training. However, these approaches primarily focus on regulating reward magnitudes or relying on expensive LLM judges, rather than verifying the logical validity of intermediate steps. Furthermore, while tree-based exploration~\cite{treepsrag} and inference-time refinement modules~\cite{smartsearch} can help mitigate errors, they introduce considerable latency and fail to internalize reasoning capabilities into the policy. Therefore, it remains challenging to provide dense, exploration-derived process supervision directly during online training.

In this paper, we propose \textbf{ProRAG}, a process-supervised reinforcement learning framework designed to resolve this credit assignment problem in the multi-hop RAG task. Unlike previous approaches that rely on sparse outcome signals, static offline preferences, or heuristic rules, ProRAG integrates learned step-level supervision directly into the online optimization loop. Specifically, we start with a Supervised Policy Warmup to initialize the model with a structured reasoning format. Leveraging this policy as a prior, we establish a fine-grained evaluator by training a Process Reward Model (PRM) on diverse reasoning paths constructed via Monte Carlo Tree Search (MCTS). To bridge the alignment gap between the initial policy and this evaluator, we introduce a Reasoning Refinement stage, which effectively warms up the model using high-quality trajectories filtered by the PRM. Finally, we implement a Process-Supervised Reinforcement Learning stage with a dual-granularity mechanism. By aggregating step-level process advantages from the PRM with global outcome rewards, ProRAG provides immediate and precise feedback for each action. This mechanism allows the model to dynamically identify and correct intermediate errors, learning how to reason correctly rather than overfitting to the final answer.

We conduct extensive experiments on five challenging multi-hop reasoning benchmarks, including PopQA~\cite{popqa}, HotpotQA~\cite{hotpotqa}, 2WikiMultihopQA~\cite{2wiki}, MuSiQue~\cite{musique}, and Bamboogle~\cite{bamboogle}. Experimental results demonstrate that ProRAG achieves superior performance compared to strong outcome-based and process-aware baselines, particularly on complex long-horizon tasks, validating that dense process supervision provides a more effective optimization signal than sparse outcome rewards or offline preference learning. Furthermore, comprehensive ablation studies confirm the essential contributions of each component in our framework, showing that the PRM-guided refinement and the dual-granularity advantage mechanism are critical for the observed performance gains. In addition, our analysis shows that ProRAG effectively regulates the retrieval process, remaining robust to irrelevant documents while adaptively planning reasoning steps according to task complexity. Efficiency evaluations further demonstrate that by internalizing reasoning capabilities, ProRAG achieves high data efficiency and low inference latency, facilitating real-world deployment.

The main contributions of this paper include:

(1) We propose ProRAG, a process-supervised reinforcement learning framework that resolves the credit assignment problem in multi-hop RAG tasks via a dual-granularity advantage mechanism combining learned step-level process rewards with outcome signals.

(2) We introduce an MCTS-based Process Reward Model that leverages tree-search exploration to quantify intermediate reasoning quality and effectively identify process hallucinations.

(3) We design a PRM-Guided Reasoning Refinement strategy to align the policy with fine-grained process preferences, ensuring stable convergence and mitigating the cold-start problem in RL.

\begin{figure*}[t]
  \centering
  \includegraphics[width=0.75\textwidth]{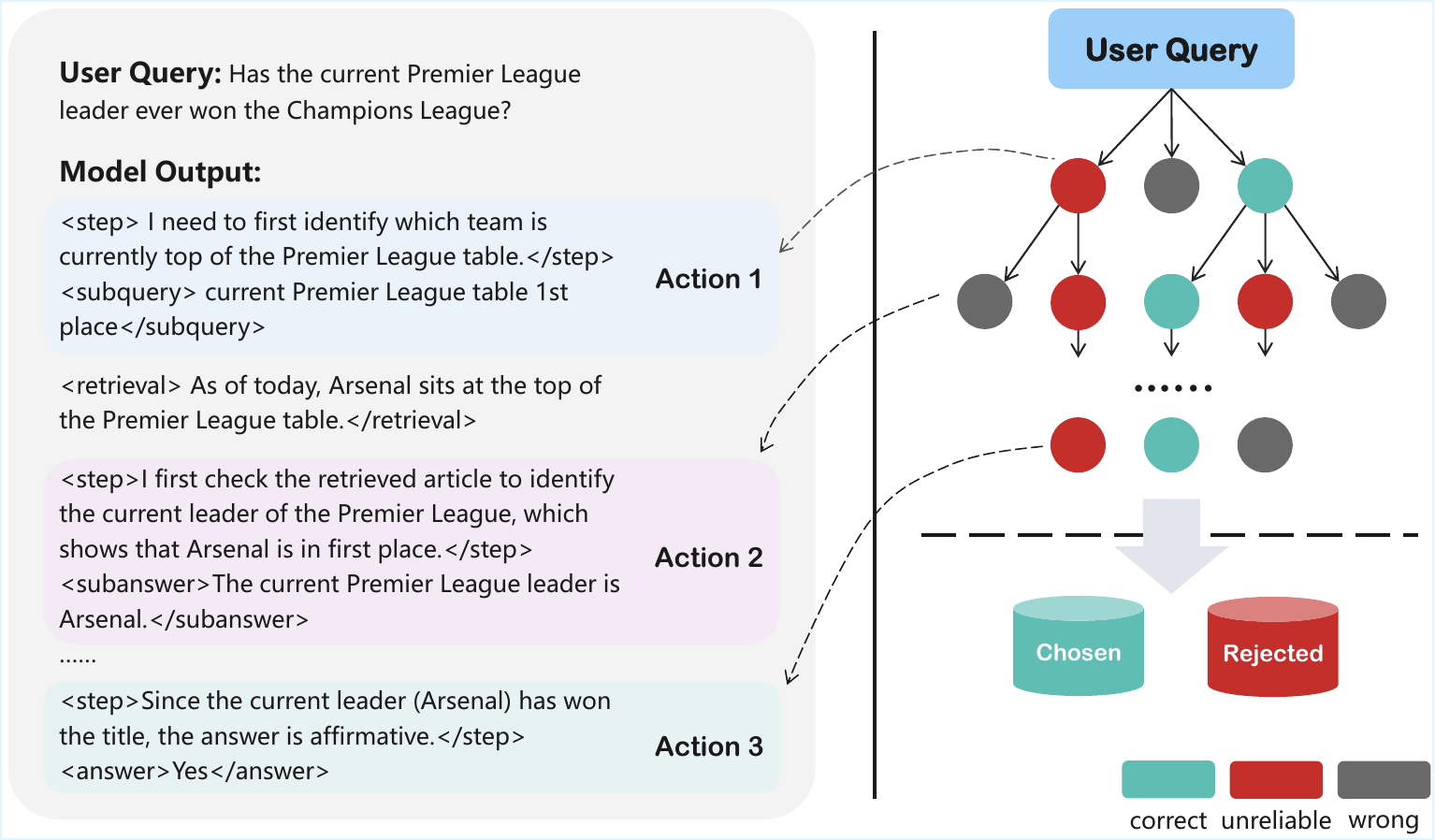}
  \caption{The left panel presents the reasoning format of our model, while the right illustrates the Monte Carlo Tree Search (MCTS) mechanism used to train the Process Reward Model.}
  \label{fig:stage1&2}
\end{figure*}

\section{Related Works}

\subsection{Retrieval-Augmented Generation}

Retrieval-Augmented Generation (RAG) has emerged as an important paradigm for mitigating hallucination and knowledge obsolescence in LLMs by dynamically incorporating relevant information from external knowledge sources~\cite{lewis2020retrieval, zhao2024retrieval}. Early research on RAG primarily focused on a standard retrieve-then-generate framework, where a retriever identifies relevant documents based on the input query, and a generator synthesizes the final answer~\cite{guu2020realm, karpukhin2020dense}. To handle more complex multi-hop queries that require multiple retrieval steps, subsequent studies have introduced iterative retrieval mechanisms. Methods such as Iter-RetGen~\cite{iter-retgen} and IRCoT~\cite{ircot} alternate between retrieval and generation steps to progressively refine the context. Furthermore, active strategies like FLARE~\cite{flare} and Self-RAG~\cite{selfrag} improve efficiency via dynamic retrieval triggered by low-confidence generation or token-level signals.

However, these pipelines typically follow predefined workflows, limiting their adaptability to handle open-ended complex reasoning tasks. To address this, recent research has explored Agentic RAG, where LLMs serve as autonomous agents that orchestrate the retrieval and reasoning process~\cite{gao2024retrieval}. In this paradigm, models explicitly plan search queries, interact with search engines, and evaluate retrieved evidence. Foundational works like ReAct~\cite{react} demonstrated the potential of interleaving reasoning traces with external tool actions. Building on this, recent advanced systems such as Search-o1~\cite{searcho1} and DeepRAG~\cite{deeprag} model the RAG process as a multi-step decision-making task to handle long-horizon queries. Despite these architectural advances, most existing agentic RAG systems rely primarily on supervised fine-tuning or complex prompting engineering, fundamentally limiting their ability to generalize to novel scenarios where optimal reasoning trajectories are unavailable.

\subsection{Reinforcement Learning for RAG}

Reinforcement learning (RL) has become a promising paradigm for optimizing RAG systems by maximizing task-specific rewards to overcome the limitations of static supervision. Recent advances, such as Search-R1~\cite{searchr1} and R1-Searcher~\cite{r1searcher}, have demonstrated that outcome-supervised RL can significantly enhance the model's ability to invoke search tools and reason over retrieved content. However, as noted in recent studies~\cite{reasonrag, decexrag}, outcome-based approaches can suffer from reward sparsity and gradient conflict in long-horizon tasks. The correct final answer does not guarantee a correct reasoning process, often leading to ``process hallucinations'' where models are reinforced for reaching the right conclusion via flawed logic or redundant retrieval~\cite{hiprag}.

To address this credit assignment problem, recent works have explored integrating process-level supervision, broadly categorized into offline and online approaches. Offline methods~\cite{reasonrag, rearter, decexrag} utilize search algorithms like MCTS or decoupled execution to synthesize high-quality preference datasets for DPO. However, these methods rely on static datasets, limiting the model's ability to adapt to on-policy distribution shifts. Conversely, online methods attempt to provide denser feedback during training. Some approaches~\cite{hiprag, ppr, stepsearch} introduce step-wise heuristic reward shaping or normalization strategies. More recent frameworks~\cite{sirag, tiresrag} leverage multi-agent interactions or multi-dimensional reward functions to guide reasoning. While effective, these methods typically rely on expensive external LLM judges or manual rule design, lacking the efficiency of a learned verifier. Furthermore, while tree-based exploration~\cite{treepsrag} and inference-time refinement modules~\cite{smartsearch} can help mitigate errors, they introduce considerable computational latency during training and inference, often failing to efficiently internalize reasoning capabilities into the policy model. In contrast, our ProRAG integrates a learned MCTS-based Process Reward Model directly into the online RL loop. By employing a dual-granularity advantage mechanism, ProRAG provides more precise, dense, and efficient supervision.

\section{ProRAG}

Existing reinforcement learning approaches for RAG often suffer from reward sparsity and inefficient credit assignment in complex multi-step reasoning. To address these challenges, we propose \textbf{ProRAG}, which introduces process rewards for step-level fine-grained optimization, thereby enhancing the reasoning capabilities of the RAG model. Our framework consists of four stages: (1) \textbf{Supervised Policy Warmup} to align the model with a multi-step reasoning format; (2) \textbf{MCTS-based Process Reward Modeling} to construct a step-level process reward model; (3) \textbf{PRM-Guided Reasoning Refinement} to rapidly enhance the model policy; and (4) \textbf{Process-Supervised Reinforcement Learning} to perform the final optimization via fine-grained rewards.

\subsection{Supervised Policy Warmup}

To equip the LLM with the capability for complex multi-hop reasoning and autonomous retrieval, we begin with a Supervised Policy Warmup. In this stage, we fine-tune the model using a constructed dataset that follows a structured reasoning–action format, establishing a stable reference policy $\pi_{\text{sft}}$ for subsequent training.

Since standard QA benchmarks typically provide sub-query chains and corresponding retrieval documents without any explicit reasoning process, we construct a high-quality dataset to teach the model this capability. To achieve this, we leverage a superior LLM (e.g., GPT-4o) to synthesize detailed chain-of-thought trajectories. By prompting the teacher model to generate the reasoning process for each step based on the provided sub-query and document, we obtain a diverse set of reasoning paths that bridge the gap between the initial query and the final answer.

To enable fine-grained process supervision, we organize the synthesized trajectories into a structured schema, as illustrated in the left panel of Figure~\ref{fig:stage1&2}. Each reasoning trajectory consists of a sequence of steps enclosed by special control tokens: \texttt{<step>} marks the internal thought process for planning; \texttt{<subquery>} denotes the generated keywords for the retrieval system; \texttt{<retrieval>} wraps the external evidence returned by the environment; \texttt{<subanswer>} indicates intermediate conclusions; and \texttt{<answer>} represents final answer. This structured format transforms unstructured reasoning into a sequence of discrete and evaluable actions.

Leveraging these structured trajectories, we perform supervised fine-tuning to establish the reference policy $\pi_{\text{SFT}}$. Rather than training on continuous sequences where the learning signal might be obscured by long reasoning process and retrieval documents, we formulate the training data as specific input-target pairs. The input comprises the historical context, including frozen retrieval content, while the target output consists of the current reasoning and action block (i.e., \texttt{<step>} and \texttt{<subquery>}). This approach directs the model’s attention entirely towards the immediate generation task.

To further enforce structural format, we adopt a format-aware training objective:
\begin{equation}
\label{eq:sft_loss}
\begin{aligned}
\mathcal{L}_{\text{SFT}} = & - \sum_{t \notin \mathcal{T}_{\text{ctrl}}} \log P(y_t \mid \bm{c}, \bm{y}_{<t}) 
    - \lambda \sum_{t \in \mathcal{T}_{\text{ctrl}}} \log P(y_t \mid \bm{c}, \bm{y}_{<t}), 
\end{aligned}
\end{equation}
where $\mathcal{T}_{\text{ctrl}}$ represents the set of special control tokens (e.g., \texttt{<step>}, \texttt{<subquery>}). By setting $\lambda > 1$, we ensure our model robustly follows the format schema in this stage.

\subsection{MCTS-based Process Reward Modeling}

Although the SFT phase equips the model with the fundamental reasoning format, only relying on final answer correctness for supervision remains insufficient. This coarse-grained feedback often leads to ``process hallucinations'', where the model arrives at a correct answer via flawed logic or redundant retrieval actions. To mitigate this, we introduce a Process Reward Model (PRM) to evaluate the validity of intermediate reasoning steps. 

To achieve this, we employ Monte Carlo Tree Search (MCTS) to construct a dataset of diverse reasoning paths with step-level labels. Specifically, we formulate the reasoning process as a tree search problem, utilizing the SFT policy $\pi_{\text{sft}}$ as a prior. Here, the state $s_t$ represents the accumulated reasoning history (including retrieval contexts), and the action $a_t$ corresponds to a generated reasoning step.

In the selection phase, we traverse the tree and \textit{select} child nodes maximizing the Predictor Upper Confidence Bound (PUCT) to balance exploitation of high-value paths and the exploration of uncertain ones:
\begin{equation}
a_t = \arg\max_{a} \left( Q(s_t, a) + c_{\text{puct}} \cdot \pi_{\text{SFT}}(a|s_t) \frac{\sqrt{\sum_{a'} N(s_t, a')}}{1 + N(s_t, a)} \right).
\end{equation}

Upon reaching a leaf node, we \textit{expand} new candidate steps by sampling candidate steps from $\pi_{sft}$ at a high temperature. We then perform \textit{simulation} to generate complete trajectories and evaluate the terminal state against the ground truth to obtain a binary correctness score $v \in \{0,1\}$, which is \textit{backpropagated} along the search path to update visit counts and mean action values. For a node corresponding to step $t \leq T$, we update its action value as:
\begin{equation}
Q(s_t, a_t) \leftarrow \frac{Q(s_t, a_t) \cdot N(s_t, a_t) + \gamma^{T-t} \cdot v}{N(s_t, a_t) + 1},
\end{equation}
where $\gamma$ is a decay factor that discounts rewards for redundant trajectories and encourages more direct solutions.

Through iterative simulations, MCTS effectively explores the solution space and accumulates a rich set of trajectories that distinguish between valid reasoning paths and those leading to errors.

While MCTS enables efficient exploration and assigns values to reasoning trajectories, the resulting Q-values remain primarily outcome-oriented. A reasoning step may receive a high value simply because it leads to a correct final answer, even if the step itself contains logical errors or unsupported assumptions. Therefore, such outcome-based signals are insufficient for reliably supervising fine-grained reasoning quality.

To obtain fine-grained process-level supervision, we further construct contrastive step-level preference pairs from the MCTS search tree. Specifically, we select sibling nodes that share the same parent context but differ at the current reasoning step, and employ GPT-4o as a logical judge to compare their reasoning quality. The model labels each pair as ``Chosen'' ($y^+$) or ``Rejected'' ($y^-$) based on strict logical validity. To validate the reliability of these labels, we manually evaluated 50 randomly sampled pairs and observed a 96\% agreement rate between GPT-4o and human annotators, indicating high label reliability. These contrastive pairs filter out noise from outcome-based signals and provide direct supervision for the PRM, enabling it to distinguish logically valid steps from flawed alternatives under identical contexts.

Finally, we train the Process Reward Model $\mathcal{R}_\phi$ by these constructed contrastive pairs, to quantify the quality of intermediate reasoning. The model takes the query context and a specific reasoning step as input and outputs a scalar score. We optimize $\mathcal{R}_\phi$ using a pairwise ranking loss to maximize the score margin between the chosen and rejected steps:
\begin{equation}
    \mathcal{L}_{\text{PRM}} = - \mathbb{E}_{(x, y^+, y^-) \sim \mathcal{D}} \left[ \log \sigma \left( \mathcal{R}_\phi(x, y^+) - \mathcal{R}_\phi(x, y^-) \right) \right],
\end{equation}
where $\sigma$ denotes the sigmoid function and $D$ represents the collected preference dataset. By minimizing this loss, $\mathcal{R}_\phi$ serves as a dense verifier to provide step-level feedback for the subsequent stages.

\subsection{PRM-Guided Reasoning Refinement}

\begin{figure*}[t]
  \centering
  \includegraphics[width=0.9\textwidth]{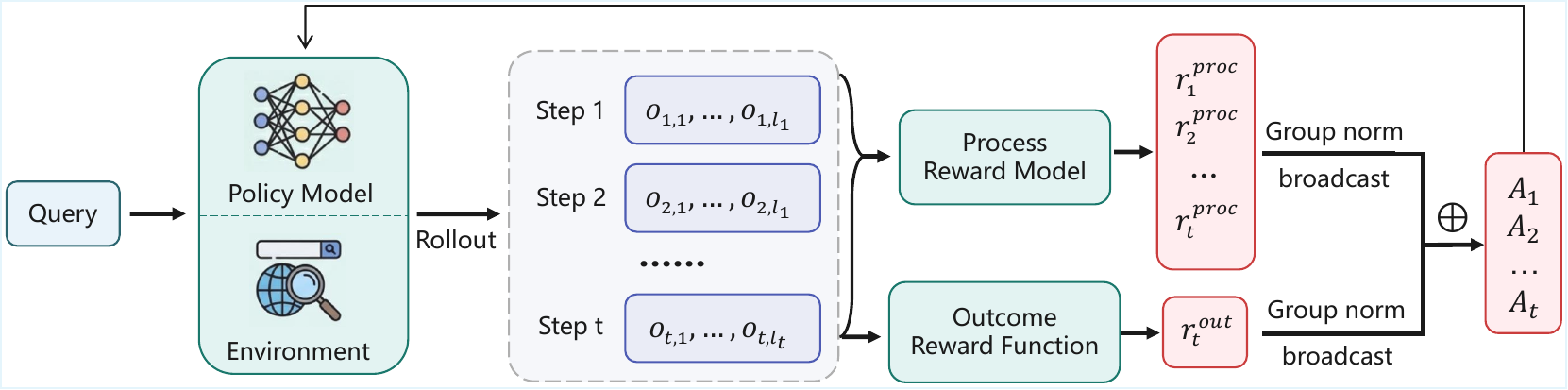}
  \caption{The framework of Process-Supervised Reinforcement Learning.}
  \label{fig:stage4}
\end{figure*}

Before the reinforcement learning phase, it is crucial to bridge the alignment gap between the initial SFT policy and the fine-grained preferences of the PRM. While $\pi_{\text{sft}}$ learns the desired structural format via SFT, its output distribution remains broad in quality, often deviating from the logical standards encoded in the PRM. Directly optimizing such an unaligned policy via RL typically suffers from severe cold-start issues due to the sparsity of high-reward trajectories in the initial search space. To address this, we propose a Reasoning Refinement stage as an intermediate step between SFT and reinforcement learning. By fine-tuning the policy model on the high-quality reasoning trajectories filtered by the PRM, we improve step-level process correctness and reduce the distribution mismatch between the SFT policy and the PRM. As a result, the refined policy provides a well-aligned and stable initialization for subsequent reinforcement learning.

To achieve this, we employ a Step-level Rejection Sampling Fine-Tuning (RFT) strategy to construct a refinement dataset. Specifically, we utilize the SFT policy $\pi_{\text{sft}}$ to generate $N$ candidate trajectories for each query. To ensure that the policy model learns only from optimal reasoning paths, we apply a dual-criterion filtering mechanism. First, we strictly require outcome correctness, retaining only those trajectories that reach the correct final answer. Second, we impose process validity by evaluating every intermediate step using our trained PRM $\mathcal{R}_\phi$. A specific (context, action) pair is preserved for training if and only if its reward score exceeds a predefined threshold (e.g., $\mathcal{R}_\phi (s_t > 0)$). This selective mechanism allows us to collect high-quality reasoning segments, thereby selective aligning the policy with the PRM's high-value regions.

Finally, we fine-tune the initial policy $\pi_{\text{SFT}}$ on these curated trajectories. By treating the (context, action) pairs as high-quality demonstrations, we optimize the model using the standard next-token prediction objective. This process effectively distills the sparse and high-value signals from the PRM into the parameters of the refined policy model $\pi_{\text{RFT}}$. As a result, $\pi_\text{rft}$ exhibits substantially stronger reasoning capability than the initial policy $\pi_{\text{sft}}$, effectively mitigating the cold-start problem in the subsequent reinforcement learning stage.

\subsection{Process-Supervised Reinforcement Learning}

Following the Reasoning Refinement stage, we employ Reinforcement Learning (RL) to further enhance the model's capabilities. However, traditional RL algorithms such as PPO~\cite{ppo} and GRPO~\cite{grpo} often rely on sparse outcome rewards and suffer from the credit assignment problem in multi-hop RAG tasks, preventing the policy model from identifying specific erroneous steps or verifying intermediate reasoning. To address this, we propose ProRAG, a reinforcement learning framework with a dual-granularity advantage mechanism. As illustrated in Figure~\ref{fig:stage4}, it aggregates group-normalized advantages from both step-level process signals and final outcome rewards, providing fine-grained and dense supervision throughout the entire reasoning process.

\subsubsection{Group Trajectory Sampling.}

To estimate policy variance, we sample groups of trajectories from the current policy $\pi_\theta$, which is initialized from the refinement model $\pi_{\text{rft}}$. Specifically, for a given query $q$, we sample a group of $G$ independent trajectories $\mathcal{Y} = \{y_1, y_2, \dots, y_G\}$ from $\pi_\theta$. Each trajectory $y_i$ is formulated as a sequence of discrete reasoning steps $y_i = (s_{i,1}, s_{i,2}, \dots, s_{i,T})$, where each step $s_{i,t} = (o_{i, t, 1}, \dots, o_{i, t, l_{i,t}})$ corresponds to a sub-sequence of tokens generated by the policy.

\subsubsection{Reward Formulation}

To evaluate these generations, we derive two distinct reward signals from dense step-level supervision and sparse trajectory-level verification.

First, we compute a \textbf{step-level process reward} $r_{i,t}^{\text{step}}$ for each intermediate reasoning step $s_{i,t}$ within a trajectory $y_i$. This reward assesses the quality of the action based on the local format constraint and the probability score from our frozen PRM $\mathcal{R}_\phi$ . Formally, given the historical context $\bm{c}_{i,t}$, the process reward is defined as:
\begin{equation}
r_{i,t}^{\text{step}} = \mathcal{R}_\phi(\bm{c}_{i,t}, s_{i,t}) + \nu_{1} \cdot \mathbb{I}_{\text{step}}(s_{i,t}),
\end{equation}
where $\mathbb{I}_{\text{step}}(\cdot)$ is an indicator function that returns $1$ if the current step $s_{i,t}$ strictly adheres to the specified tag schema (e.g., correct usage of \texttt{<step>} or \texttt{<subquery>}) and $0$ otherwise. $\nu_{1}$ is a coefficient for this format bonus.

Second, we assign a terminal \textbf{outcome reward} $r_{i}^{\text{out}}$ after the completion of the trajectory to evaluate the global utility. This reward is determined by the correctness of the final answer and the format integrity of the complete reasoning chain:
\begin{equation}
r_{i}^{\text{out}} = \text{F1}(\hat{a}_i, a^*) + \nu_{2} \cdot \mathbb{I}_{\text{traj}}(y_i).
\end{equation}
Here, $\hat{a}_i$ denotes the predicted answer extracted from $y_i$, and $a^*$ is the ground truth. The function $\text{F1}(\cdot)$ computes the token-level accuracy between the prediction and the ground truth. The indicator function $\mathbb{I}_{\text{traj}}(y_i)$ evaluates whether the entire trajectory $y_i$ follows the complete reasoning-action workflow, discouraging the policy from bypassing required retrieval or intermediate reasoning steps. $\nu_{2}$ is a coefficient for this format bonus.

\subsubsection{Dual-Granularity Advantage Estimation.}

To address the credit assignment problem in multi-turn trajectories, we propose a dual-granularity advantage estimation strategy, which aggregates dense step-level supervision and sparse outcome reward.

For a generated trajectory $y_i$, we broadcast the rewards to the token level and define the \textbf{process advantage} $A_{i,t,k}^{\text{proc}}$ and \textbf{outcome advantage} $A_{i,t,k}^{\text{out}}$ for the $k$-th token within step $s_{i,t}$ as follows:
\begin{equation}
A_{i,t,k}^{\text{proc}} = \frac{r_{i,t}^{\text{step}} - \mu^{\text{step}}}{\sigma^{\text{step}}}, \quad
A_{i,t,k}^{\text{out}} = \frac{r_{i}^{\text{out}} - \mu^{\text{out}}}{\sigma^{\text{out}}},
\end{equation}
where $\mu$ and $\sigma$ denote the group-wise mean and standard deviation of the corresponding rewards. By this mechanism, the reward signals are broadcast to all tokens $k$ within each step $s_{i,t}$, providing dense supervision for optimization.

By weighted aggregation of the process advantage and the outcome advantage, we obtain the \textbf{total advantage} $A_{i,t,k}$ as:
\begin{equation}
A_{i,t,k} = A_{i,t,k}^{\text{out}} + \beta \cdot A_{i,t,k}^{\text{proc}}.
\label{adv}
\end{equation}

This dual-granularity advantage estimation mitigates the limitations of single-source supervision. Using only sparse outcome rewards provides insufficient credit assignment for intermediate reasoning steps, while relying solely on step-level process rewards may encourage optimization of local signals at the expense of global correctness. By combining outcome advantages with weighted process advantages, ProRAG delivers dense token-level supervision while preserving alignment with the final reasoning objective.

\subsubsection{Policy Optimization.}

Finally, we optimize the policy $\pi_\theta$ by maximizing the total advantages over the trajectory groups. Consistent with GRPO~\cite{grpo}, we compute baselines via the group statistics in Eq.~\ref{adv} instead of a value network, ensuring memory efficiency. The loss function is formulated as:
\begin{equation}
\begin{aligned}
\mathcal{L}(\theta) = -\mathbb{E}_{q \sim \mathcal{D}, \mathcal{Y} \sim \pi_{\text{old}}} \bigg[ \frac{1}{G} & \sum_{i=1}^G \sum_{t=1}^{T_i} \sum_{k=1}^{l_{i,t}} \min \Big( \rho_{i,t,k} A_{i,t,k}, \\
& \text{clip}(\rho_{i,t,k}, 1-\varepsilon, 1+\varepsilon) A_{i,t,k} \Big) \bigg],
\end{aligned}
\end{equation}
where $\rho_{i,t,k} = \frac{\pi_\theta(o_{i,t,k} \mid \bm{h}_{i,t,k})}{\pi_{\text{old}}(o_{i,t,k} \mid \bm{h}_{i,t,k})}$ represents the probability ratio between the current and reference policies, where $\bm{h}_{i,t,k}$ includes all context preceding token $o_{i,t,k}$ and $\varepsilon$ is the clipping coefficient.

\section{Experiment}

\subsection{Dataset and Metrics}

We evaluate ProRAG and all baseline models on five diverse benchmarks, including one general QA dataset PopQA~\cite{popqa}, and four multi-hop QA datasets: HotpotQA~\cite{hotpotqa}, 2WikiMultiHopQA~\cite{2wiki}, MuSiQue~\cite{musique}, and Bamboogle~\cite{bamboogle}. These datasets encompass a wide spectrum of challenges, from long-tail knowledge retrieval to complex multi-step deductive reasoning, providing a comprehensive evaluation of the models' ability to plan, retrieve, and reason across varying levels of difficulty and distribution shifts. For these benchmarks, we report Exact Match (EM) and F1 Score as metrics to evaluate the answer accuracy. Furthermore, to verify the statistical significance of our performance gains, we conduct a paired t-test comparing ProRAG against the strongest baseline with a significance level of $p<0.05$.

\subsection{Baselines}

We compare ProRAG against a comprehensive set of baselines, which can be categorized into three types as follows:

(1) \textbf{Standard Baselines.} These include Na\"ive Generation and Standard RAG, which represent the lower bound of performance. Na\"ive Generation relies solely on the parametric knowledge of the pre-trained LLM without external retrieval. Standard RAG~\cite{lewis2020retrieval} represents the traditional retrieve-then-generate framework, which performs a single-step retrieval based on the initial query before generating the final answer.

(2) \textbf{Advanced Baselines.} This category encompasses inference-time methods that dynamically adjust the search process or employ agentic planning. We evaluate Iter-RetGen\cite{iter-retgen} and IRCoT\cite{ircot}, which interleave retrieval and reasoning steps in an iterative loop to progressively gather evidence. Additionally, we include FLARE\cite{flare}, an active retrieval strategy that triggers search actions based on generation confidence. Furthermore, we benchmark against Search-o1\cite{searcho1}, a strong agentic baseline that treats LLMs as autonomous agents capable of multi-step planning, task decomposition, and reflection to explicitly control the retrieval workflow.

(3) \textbf{Reinforcement Learning Baselines.} Finally, we compare against state-of-the-art methods that optimize the policy through reinforcement learning. For outcome-based RL, we select \textbf{Search-R1}~\cite{searchr1} as the primary baseline, which employs PPO to optimize the policy based on the final answer. For offline process supervision, we include \textbf{ReasonRAG}~\cite{reasonrag}, which uses MCTS to synthesize static preference data for DPO, providing step-level signals for optimization. For online process-aware RL, we compare with HiPRAG~\cite{hiprag}, a recent method that shapes rewards via sub-goal heuristics.

\subsection{Implementation Details}

We implement our framework based on the HuggingFace Transformers~\cite{transformers} and TRL~\cite{trl} Libraries, with Qwen3-8B~\cite{qwen3} serving as the backbone for both our method and all baselines. For the retrieval environment, we use the 2018 Wikipedia dump~\cite{wikicorpus} as the knowledge source and E5-base~\cite{e5} as the retriever, retrieving the top 3 documents for each query.

We conduct all experiments on 4 NVIDIA A100 (80GB) GPUs. For the Supervised Policy Warmup stage, we perform full-parameter fine-tuning on 109k MuSiQue context–action pairs for one epoch to initialize the model’s reasoning capability. In the Process Reward Modeling stage, we sample 728 queries from HotpotQA and MuSiQue, and then perform MCTS with 200 simulations, an expansion width of $K=5$, a maximum depth of 10, an exploration constant $c_{\text{puct}}=2.5$, and a discount factor $\gamma=0.99$. For these trees, we collect 8,255 contrastive pairs labeled by GPT-4o, and train the scalar regression head from the SFT checkpoint for three epochs. Next, in the Reasoning Refinement Stage we select 105k high-quality trajectories from the two datasets filtered by the trained PRM, and fine-tune the policy model for one additional epoch to enhance reasoning ability. Finally, in the Process-Supervised RL stage, we apply LoRA fine-tuning to optimize the policy on 10k held-out queries drawn from both datasets for 1 epoch, with a generation group size of $G=8$ and a dual-granularity weight $\beta=0.3$.

\begin{table*}[t]
\centering
\caption{Overall results of ProRAG and baselines on five benchmarks. The best and second-best scores are highlighted in \textbf{bold} and \underline{underlined}. $^\dagger$ indicates statistically significant improvements over the best baseline (Search-R1) with $p < 0.05$.}
\label{tab:main_results}
\setlength{\tabcolsep}{5pt}
\begin{tabular}{>{\raggedright\arraybackslash}p{0.16\linewidth} cc cc cc cc cc cc}
\toprule
\multirow{2}{*}{\textbf{Method}} & \multicolumn{2}{c}{\textbf{PopQA}} & \multicolumn{2}{c}{\textbf{HotpotQA}} & \multicolumn{2}{c}{\textbf{2Wiki}} & \multicolumn{2}{c}{\textbf{MuSiQue}} & \multicolumn{2}{c}{\textbf{Bamboogle}} & \multicolumn{2}{c}{\textbf{Avg.}} \\
\cmidrule(lr){2-3} \cmidrule(lr){4-5} \cmidrule(lr){6-7} \cmidrule(lr){8-9} \cmidrule(lr){10-11} \cmidrule(lr){12-13}
& EM & F1 & EM & F1 & EM & F1 & EM & F1 & EM & F1 & EM & F1 \\
\midrule
\multicolumn{13}{l}{\textbf{Standard Methods}} \\
Na\"ive Generation & 11.7 & 16.6 & 9.8 & 19.0 & 15.8 & 21.0 & 2.1 & 9.5 & 4.8 & 11.0 & 8.8 & 15.4 \\
Standard RAG & 27.0 & 35.9 & 20.0 & 31.5 & 16.1 & 23.0 & 3.3 & 10.3 & 12.0 & 20.1 & 15.7 & 23.3 \\
\midrule
\multicolumn{13}{l}{\textbf{Advanced Methods}} \\
Iter-RetGen & 32.9 & 37.8 & 15.9 & 21.2 & 9.3 & 11.5 & 4.8 & 8.5 & 17.6 & 23.7 & 16.1 & 20.5 \\
IRCoT & 33.6 & 42.2 & 27.2 & 37.9 & 25.1 & 30.4 & 5.1 & 13.1 & 16.8 & 28.6 & 21.6 & 30.4 \\
FLARE & 39.3 & 46.3 & 28.9 & 40.1 & 24.1 & 29.5 & 5.8 & 13.5 & 27.2 & 37.8 & 25.1 & 33.4 \\
Search-o1 & 38.4 & 43.9 & 28.6 & 38.1 & 22.6 & 27.9 & 13.0 & 19.9 & 38.4 & 50.1 & 28.2 & 36.0 \\
\midrule
\multicolumn{13}{l}{\textbf{Reinforcement Learning-based Methods}} \\
Search-R1 & \underline{45.9} & \underline{50.1} & \textbf{42.7} & \textbf{54.6} & 39.9 & 45.2 & \underline{20.6} & \underline{29.5} & \underline{43.2} & \underline{54.1} & \underline{38.5} & \underline{46.7} \\
ReasonRAG & 41.5 & 46.2	& 38.4 & 48.9 & \underline{43.6} & \underline{50.4} & 12.8 & 20.6 & 36.0 & 45.5 & 34.5 & 42.3\\
HiPRAG & 34.3 & 42.7 & 36.8 & 49.1 & 37.0 & 43.2 & 15.6 & 25.0 & 39.2 & 51.6 &  32.6 & 42.3 \\
\textbf{ProRAG (Ours)} & \textbf{47.2}$^\dagger$ & \textbf{51.6}$^\dagger$ & \underline{41.4} & \underline{52.8} & \textbf{46.0}$^\dagger$ & \textbf{51.1}$^\dagger$ & \textbf{23.5}$^\dagger$ & \textbf{34.1}$^\dagger$ & \textbf{45.6} & \textbf{56.4} & \textbf{40.7}$^\dagger$ & \textbf{49.2}$^\dagger$ \\
\bottomrule \\
\end{tabular}

\end{table*}

\subsection{Main Results}

The overall performance of ProRAG and baseline models across five benchmarks is presented in Table~\ref{tab:main_results}. 

\textbf{Comparison with Standard and Advanced Baselines.} 
As shown in the table, standard and advanced baselines generally struggle to achieve high accuracy compared to RL-based methods. Specifically, Na\"ive Generation performs worst due to internal knowledge limitations, while Standard RAG provides only modest improvements, suggesting that single-step retrieval is insufficient for complex multi-hop dependencies. Although iterative strategies like Iter-RetGen and IRCoT improve upon standard methods by interleaving retrieval and reasoning, their performance gains are limited by the accumulation of noise from retrieved documents. Similarly, while dynamic methods such as FLARE and agentic baselines like Search-o1 achieve competitive results on specific tasks (e.g., Bamboogle), they still fall short of the optimal performance. Search-o1, for instance, is constrained by general-purpose prompting and lacks task-specific policy optimization. In contrast, reinforcement learning baselines consistently outperform these approaches across all metrics, highlighting the necessity of explicitly aligning the retrieval and reasoning policy.

\textbf{Comparison with Reinforcement Learning Baselines.}
For the RL methods, ProRAG demonstrates superior overall performance, surpassing the strongest baseline by 2.5\% in average F1 score. We observe that outcome-based approaches like Search-R1 remain competitive on HotpotQA, primarily because this benchmark serves as a standard in-domain task where the reasoning chains are relatively shallow and easier to fit using sparse rewards. However, this advantage disappears on more complex tasks. As shown in Table~\ref{tab:main_results}, ProRAG achieves statistically significant improvements ($p < 0.05$) on challenging datasets like MuSiQue and 2WikiMultihopQA, confirming that sparse outcome signals are insufficient for learning long-horizon dependencies. Similarly, offline optimization methods like ReasonRAG lag behind, as its optimization on static data lacks the on-policy exploration required to discover novel solution paths. In contrast, by integrating a learned Process Reward Model with online optimization, ProRAG provides dense feedback that effectively navigates the complex search space, leading to the highest average accuracy.

\textbf{Generalization and Robustness on Complex Tasks.}
Beyond overall averages, ProRAG exhibits superior generalization and robustness across different benchmarks. On out-of-domain datasets such as PopQA and 2WikiMultihopQA, ProRAG consistently outperforms all baseline methods, indicating that our process-level supervision fosters transferable reasoning skills rather than merely overfitting to the training distribution. Furthermore, we observe distinct performance trends on tasks requiring deeper reasoning. While baselines tend to plateau on the complex MuSiQue dataset, ProRAG maintains a clear lead (+4.6\% F1 over Search-R1), validating the necessity of step-level verification for long reasoning chains. We also note that on Bamboogle, ProRAG achieves a substantial absolute improvement of 2.3\% F1; however, due to the extremely small test set size (125 examples), this dataset provides insufficient statistical power to establish significance, despite the evident performance margin. Overall, these results demonstrate that ProRAG is the most robust framework, ensuring both in-domain stability and reliable out-of-domain generalization.

\begin{table*}[t]
\centering
\caption{Ablation analysis of different framework components. \textbf{w/o Refinement} removes the PRM-guided warmup stage; \textbf{w/o RL} relies solely on supervised fine-tuning on high-quality filtered data; \textbf{GRPO Baseline} utilizes only outcome rewards for optimization. \textbf{ProRAG (Full)} integrates all components to achieve optimal performance.}
\label{tab:ablation_full}
\setlength{\tabcolsep}{5pt}
\begin{tabular}{>{\raggedright\arraybackslash}p{0.16\linewidth} cc cc cc cc cc cc}
\toprule
\multirow{2}{*}{\textbf{Method}} & \multicolumn{2}{c}{\textbf{PopQA}} & \multicolumn{2}{c}{\textbf{HotpotQA}} & \multicolumn{2}{c}{\textbf{2Wiki}} & \multicolumn{2}{c}{\textbf{MuSiQue}} & \multicolumn{2}{c}{\textbf{Bamboogle}} & \multicolumn{2}{c}{\textbf{Avg.}} \\
\cmidrule(lr){2-3} \cmidrule(lr){4-5} \cmidrule(lr){6-7} \cmidrule(lr){8-9} \cmidrule(lr){10-11} \cmidrule(lr){12-13}
& EM & F1 & EM & F1 & EM & F1 & EM & F1 & EM & F1 & EM & F1 \\
\midrule
\textbf{ProRAG (Ours)} & 47.2 & 51.6 & \textbf{41.4} & \textbf{52.8} & \textbf{46.0} & \textbf{51.1} & \textbf{23.5}& \textbf{34.1} & \textbf{45.6} & \textbf{56.4} & \textbf{40.7} & \textbf{49.2} \\
\quad w/o Refinement & \textbf{47.3} & \textbf{51.7} & 38.0 & 49.7 & 43.2 & 48.6 & 23.2 & 32.8 & 39.2 & 52.2 & 38.2 & 47.0 \\
\quad w/o RL & 32.1 & 36.9 & 36.8 & 48.3 & 38.9 & 44.4 & 23.6 & 33.0 & 39.2 & 48.7 & 34.1 & 42.3 \\
\quad SFT Policy & 27.9 & 32.7 & 28.8 & 38.8 & 29.0 & 33.3 & 23.3 & 32.5 & 36.0 & 46.8 & 29.0 & 36.8 \\
\midrule
GRPO Baseline & 30.2 & 34.8 & 33.9 & 44.8 & 38.6 & 43.1 & 23.5 & 32.4 & 38.4 & 49.9 & 33.6 & 42.6 \\
\bottomrule \\
\end{tabular}

\end{table*}

\subsection{Ablation Studies}

To validate the effectiveness of each component in ProRAG, we conduct ablation studies by comparing our full model against four variants. We define these variants as follows: (1) \textbf{SFT Policy}: The base model fine-tuned only on the supervised warmup data; (2) \textbf{GRPO Baseline}: A strong outcome-based RL baseline that optimizes the SFT policy using Online GRPO with only sparse final rewards; (3) \textbf{w/o RL}: The model derived from the Reasoning Refinement stage, which is fine-tuned on high-quality PRM-filtered trajectories but lacks subsequent reinforcement learning; (4) \textbf{w/o Refinement}: A variant that skips the Reasoning Refinement stage and directly applies Process-Supervised RL to the SFT policy.

The experimental results are summarized in Table~\ref{tab:ablation_full}. We analyze the impact of each module as follows.

\begin{figure}[t]
  \centering
  \includegraphics[width=\columnwidth]{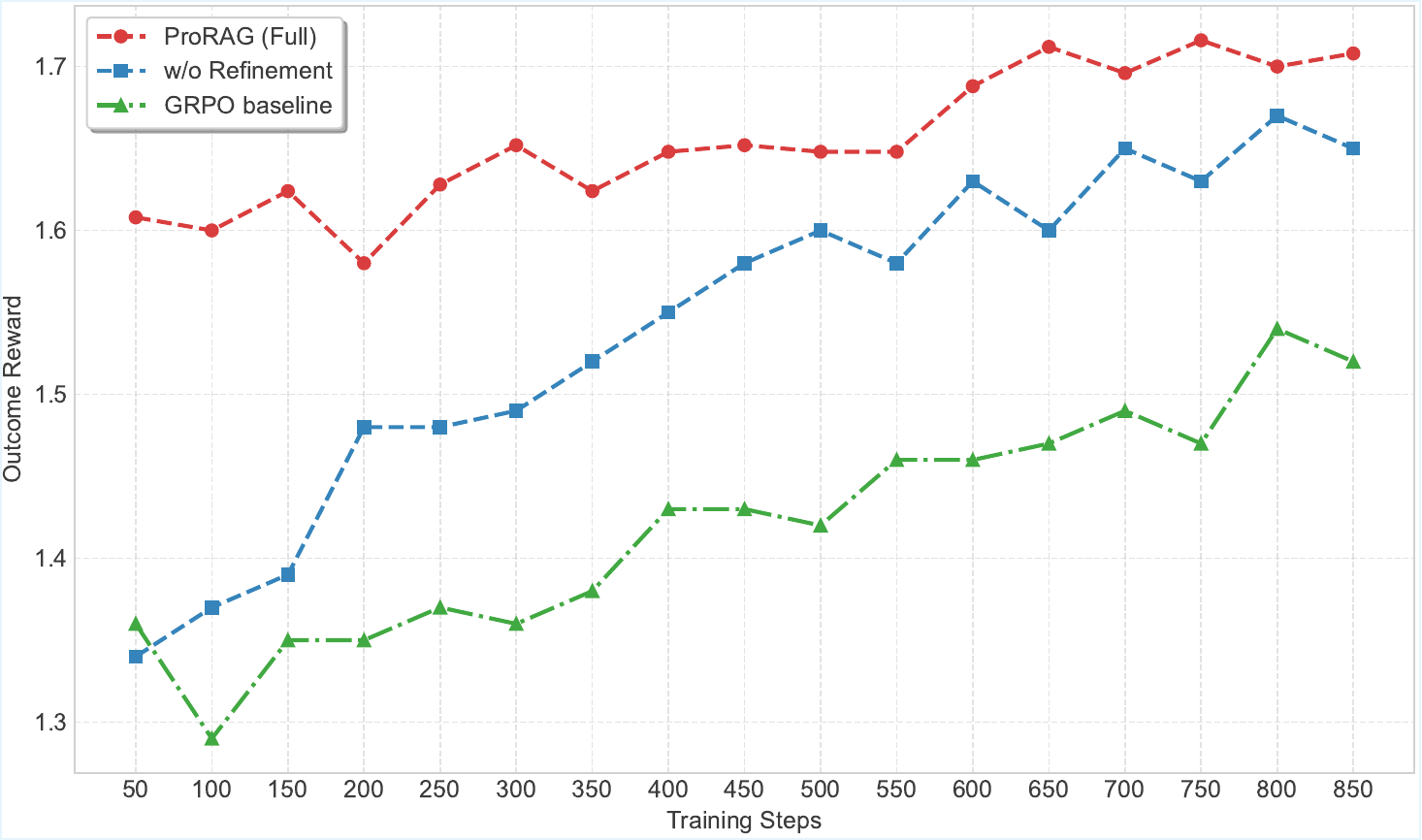}
  \caption{Training curves of ProRAG compared to the GRPO baseline and the unrefined variant (w/o Refinement).}
  \label{fig:learning_curve}
\end{figure}

\textbf{Effectiveness of Process-Supervised RL.}
Comparing ProRAG with the \textit{GRPO Baseline} in Table~\ref{tab:ablation_full}, we observe a significant and consistent performance improvement across all datasets. This advantage is further corroborated by the learning curves presented in Figure~\ref{fig:learning_curve}, where ProRAG not only exhibits a markedly faster convergence rate but also achieves a substantially higher asymptotic reward compared to the outcome-based baseline. These results strongly validate that our proposed dual-granularity advantage mechanism effectively addresses the credit assignment problem in complex multi-step reasoning. By introducing dense, step-level supervision, ProRAG enables the model to identify and reinforce correct intermediate reasoning steps, thereby leading to more efficient and stable optimization than approaches that rely solely on sparse outcome signals.

\textbf{Importance of Reasoning Refinement.}
The comparison between ProRAG and the varint \textit{w/o Refinement} demonstrates the importance of the reasoning refinement stage. Although directly applying RL to the SFT policy can lead to competitive results, the full model achieves higher overall accuracy and stability. As shown in the learning curves in Figure~\ref{fig:learning_curve}, the \textit{w/o Refinement} variant initiates with a significantly lower reward and consistently trails behind the full model throughout training, ultimately converging to a suboptimal solution. This indicates that directly exposing the SFT policy to RL leads to a misalignment between the policy's initial distribution and the PRM's preference landscape. The Refinement stage effectively bridges this gap by warming up the policy on high-quality, PRM-filtered trajectories, thereby mitigating the cold-start problem and ensuring that the subsequent RL stage begins from a region of higher reasoning validity.

\textbf{Necessity of Online Exploration.}
Finally, when comparing ProRAG with the \textit{w/o RL} variant, we observe that removing the online RL stage leads to a sharp and consistent performance drop across multiple benchmarks. While the PRM-guided Refinement effectively distills high-quality reasoning patterns from static data, it remains a  supervised process constrained by the distribution of pre-collected samples. Consequently, the model lacks the capability to generalize to novel scenarios. The subsequent Process-Supervised RL stage addresses this limitation by enabling on-policy exploration, allowing the model to actively interact with the environment and discover higher-reward reasoning trajectories that extend beyond the boundaries of the offline dataset.

\subsection{Impact of Retrieval}

In this subsection, we investigate the performance of ProRAG under different numbers of retrieved documents and retrieval steps to evaluate its context utilization and planning adaptivity.

\subsubsection{Impact of the Number of Retrieved Documents}

\begin{figure}[t]
  \centering
  \includegraphics[width=\columnwidth]{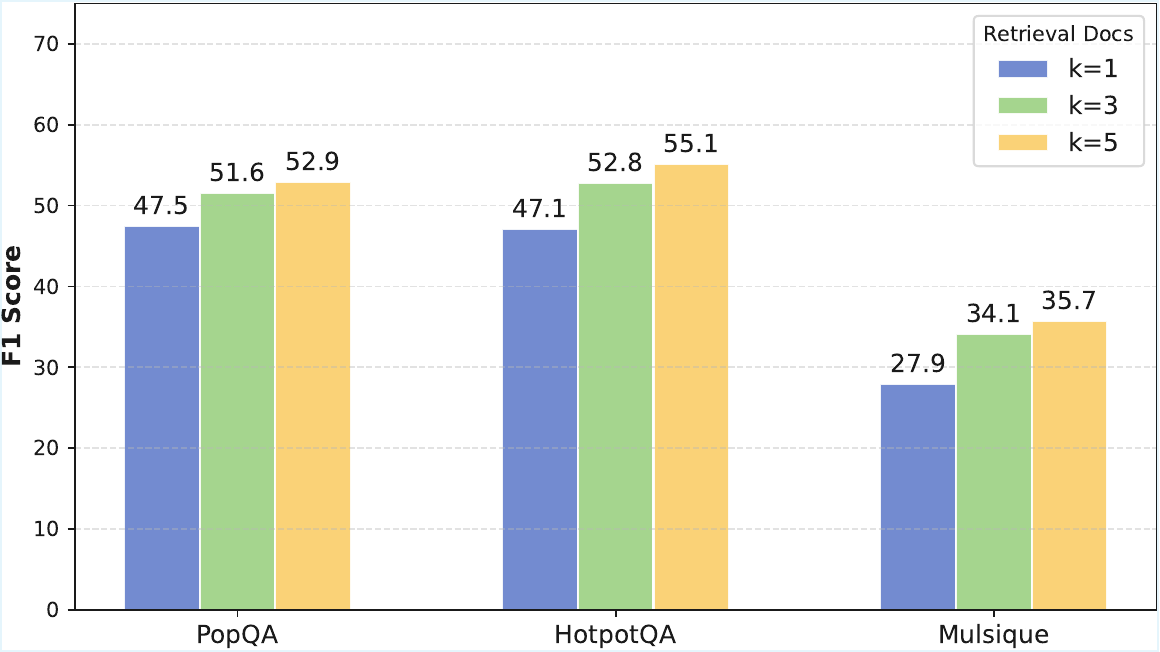}
  \caption{Performance comparison (F1 Score) on PopQA, HotpotQA, and Musique datasets with different numbers of retrieved documents ($k \in \{1, 3, 5\}$).}
  \label{fig:retrieval_k}
\end{figure}

As shown in Figure~\ref{fig:retrieval_k}, the performance gain from increasing the retrieval context ($k$) correlates strongly with task complexity. On the single-hop PopQA dataset, the performance curve saturates early as $k$ increases, indicating that a single relevant document is typically sufficient for factual recall. Notably, the model’s performance does not decrease at $k=5$, suggesting strong robustness against noise from irrelevant documents. In contrast, the complex multi-hop benchmark MuSiQue demonstrates significant sensitivity to context size, with the F1 score jumping by 6.2 points when $k$ increases to 3. This finding suggests that a restricted context window ($k=1$) creates an information bottleneck for reasoning. By providing multiple documents, ProRAG can effectively identify the key evidence, which is crucial for resolving long-horizon dependencies.

\subsubsection{Impact of the Number of Retrieval Steps}

\begin{figure}[t]
  \centering
  \includegraphics[width=\columnwidth]{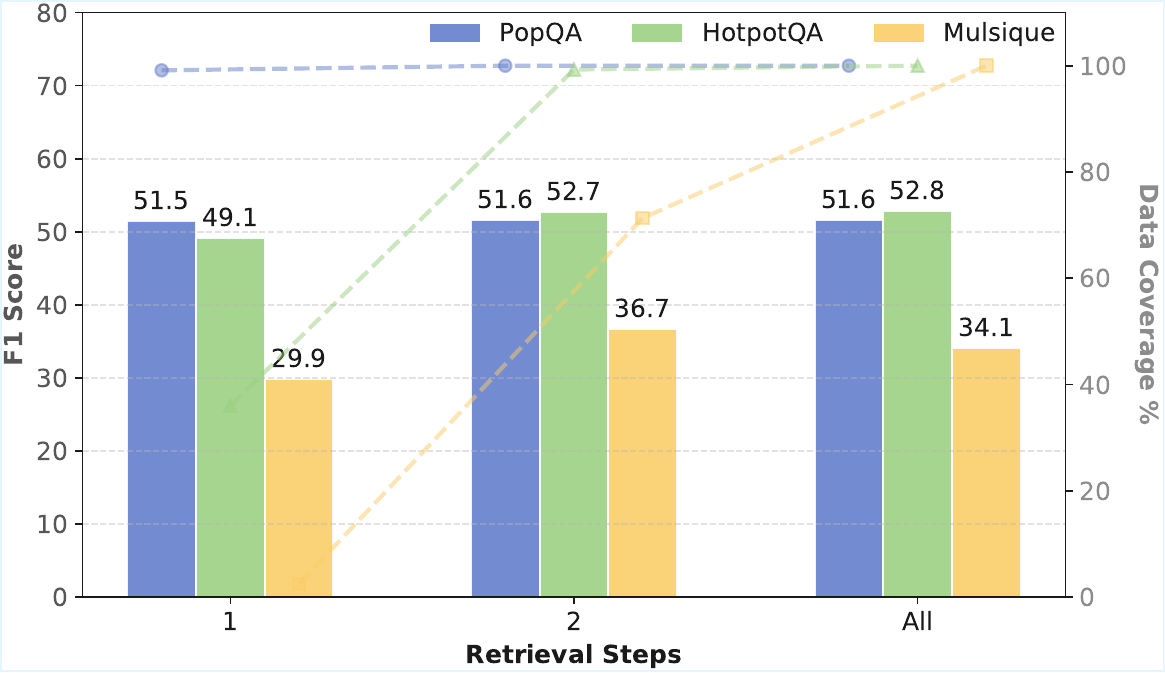}
  \caption{Cumulative Average F1 Score (bars) and Data Coverage (dashed lines) across different retrieval steps ($\leq 1$, $\leq 2$, All) on PopQA, HotpotQA, and Musique datasets.}
  \label{fig:retrieval_n}
\end{figure}

Figure~\ref{fig:retrieval_n} illustrates the cumulative average F1 score and data coverage across retrieval steps, highlighting the dynamic planning capability of ProRAG. The data coverage curves reveal distinct adaptivity patterns, where nearly 100\% of PopQA queries are efficiently resolved within the first step, thereby minimizing unnecessary computational for simple tasks. Conversely, the model autonomously extends the reasoning chain for complex benchmarks such as HotpotQA and MuSiQue, which often requires more than single step. Furthermore, while longer reasoning chains in typical iterative frameworks often suffer from performance degradation due to error propagation, ProRAG maintains robust accuracy with cumulative F1 scores increasing as retrieval steps grow. For instance, the F1 score on MuSiQue increases from 29.9 to 34.1, demonstrating that process-level supervision helps guide the model search in complex spaces and stabilize long-horizon reasoning.

\subsection{Sensitivity Analysis}

We investigate the sensitivity of model performance to the dual-granularity weight $\beta$. As shown in Table~\ref{tab:sensitivity}, performance exhibits an inverted-U trend, peaking at $\beta=0.3$. At $\beta=0.0$, the model relies solely on sparse outcome signals, struggling to identify the specific reasoning steps that lead to the correct answer, which results in the lowest performance. In contrast, introducing process supervision ($\beta > 0$) significantly boosts performance, with $\beta=0.3$ achieving the optimal results (49.2\% F1). This indicates that the dense feedback from the PRM serves as a critical guide to stabilize the reasoning trajectory and resolve the credit assignment problem. However, as $\beta$ increases further ($\beta \ge 0.3$), we observe a consistent decline. This suggests that although the PRM provides dense signals, it inevitably contains estimation errors. Consequently, an overweighted PRM causes the model to overfit to noisy step-level signals rather than the ground-truth objective, degrading the final reasoning accuracy.

\subsection{Efficiency Analysis}
In this subsection, we analyze ProRAG from two perspectives: data efficiency in training and computational cost in inference.

\subsubsection{Data Efficiency in Training.}
Standard outcome-based RL methods typically suffer from reward sparsity, requiring extensive training (e.g., \textbf{90k queries} for Search-R1) to learn reasoning patterns. In contrast, ProRAG achieves superior performance using only \textbf{10k queries} for the RL stage, significantly reducing the required training data. Furthermore, compared to offline process-optimization methods that rely on computationally expensive MCTS over the entire training set to construct preference pairs (e.g., \textbf{5k queries} for ReasonRAG), ProRAG exhibits substantially higher annotation and optimization efficiency. By performing MCTS on only \textbf{728 seed queries}, we train a generalized PRM that serves as a dense and reliable evaluator for online policy optimization, thereby avoiding expensive tree-search construction.

\subsubsection{Computational Cost in Inference.}
Recent approaches often trade inference latency for performance improvement by employing intensive test-time compute strategies, such as tree-based exploration or iterative generation-evaluation-refinement loops. While effective, these methods significantly increase the inference latency and system complexity due to multiple model invocations and trajectory regeneration. In contrast, ProRAG follows the principle of policy internalization. By incorporating process-level feedback in training, the model learns to reason and plan effectively. During inference, ProRAG generates high-quality reasoning trajectories in a single pass without external judges or iterative regeneration. This design ensures that ProRAG preserves the inference efficiency of standard LLMs, making it a more practical solution for latency-sensitive deployment.

\begin{table}[t]
\centering
\caption{Sensitivity analysis of $\beta$ on the average performance across all benchmarks. A balanced weight of $\beta=0.3$ yields the best results for both EM and F1.}
\label{tab:sensitivity}
\setlength{\tabcolsep}{6pt}

\begin{tabular}{p{0.16\columnwidth} cccccc}
\toprule
\multirow{2}{*}{\textbf{Metric}} & \multicolumn{6}{c}{\textbf{Dual-Granularity Weight} $\beta$} \\
\cmidrule(lr){2-7}
& 0.0 & 0.1 & \textbf{0.3} & 0.5 & 0.7 & 0.9\\
\midrule
EM & 38.3 & 40.3 & \textbf{40.7} & 39.2 & 38.5 & 38.4 \\
F1 score & 46.7 & 48.3 & \textbf{49.2} & 47.7 & 47.1 & 46.8 \\
\bottomrule \\
\end{tabular}

\end{table}

\section{Conclusion}

In this paper, we introduce ProRAG, a novel process-supervised reinforcement learning framework designed to resolve the credit assignment problem in multi-hop RAG tasks. Unlike traditional approaches that rely on sparse outcome signals, static offline preferences, or heuristic rules, ProRAG integrates learned, step-level supervision directly into the online optimization loop. This mechanism effectively mitigates ``process hallucinations,'' where models arrive at correct answers via flawed reasoning or redundant retrieval. By leveraging Monte Carlo Tree Search to construct a high-quality Process Reward Model and employing a dual-granularity advantage mechanism, our framework provides precise feedback for every intermediate action, enabling the model to distinguish between valid logical deductions and spurious correlations. Comprehensive experiments across five diverse benchmarks demonstrate that ProRAG achieves superior overall performance, particularly on complex long-horizon reasoning tasks. Furthermore, our analysis highlights that ProRAG achieves these gains with high data efficiency and preserves low inference latency by internalizing reasoning capabilities into the policy, making it efficient for real-world deployment. Future work will explore extending this process-supervision paradigm to more dynamic, open-ended environments to further advance the reliability of autonomous systems.


\bibliographystyle{ACM-Reference-Format}
\bibliography{sample-base}

\appendix

\end{document}